\newcommand{\CLASSINPUTtoptextmargin}{60pt} 
\newcommand{\CLASSINPUTbottomtextmargin}{43pt}

\documentclass[twocolumn,journal,10pt]{IEEEtran}
                               
\usepackage{stfloats}   
\usepackage[font=footnotesize]{caption} 
\usepackage{cite}
\usepackage[font=footnotesize]{subfig}
\usepackage{multirow}
\usepackage{ifpdf}
\usepackage[pdftex]{graphicx}
\usepackage{epstopdf}
\usepackage[cmex10]{amsmath}
\usepackage{comment}
\usepackage{array}
\usepackage{mdwmath}
\usepackage{mdwtab}
\usepackage{eqparbox}
\usepackage{verbatim}

\usepackage{amsmath}
\usepackage{amssymb}
\usepackage{algorithm}

\usepackage{algorithmicx}
\usepackage{algpseudocode}

\usepackage[font=footnotesize]{subfig}
\usepackage{fixltx2e}
\usepackage{stfloats}

\usepackage{amsmath,bm}
\usepackage{amssymb}
\usepackage{color}
\usepackage{subfig}

\usepackage{multirow}
\usepackage{float}
\usepackage{caption}

\usepackage{comment}
\usepackage{tabularx} 
\usepackage{makecell}
\usepackage{amsmath}
\usepackage{graphicx}
\usepackage[colorlinks=true, allcolors=blue]{hyperref}
\usepackage{adjustbox}
\usepackage{blindtext}

\usepackage[font=footnotesize]{subfig}
\usepackage{fixltx2e}
\usepackage{stfloats}

\usepackage{amsmath,bm}
\usepackage{amssymb}
\usepackage{color}
\usepackage{subfig}

\usepackage{multirow}
\usepackage{float}
\usepackage{caption}
\usepackage[normalem]{ulem}
\usepackage{comment}

\usepackage{cite}
\usepackage[font=footnotesize]{subfig}

\usepackage{ifpdf}

\usepackage{epstopdf}
\usepackage[cmex10]{amsmath}

\usepackage{array}
\usepackage{mdwmath}
\usepackage{mdwtab}
\usepackage{eqparbox}
\usepackage{verbatim}
\usepackage{booktabs}
\usepackage{multirow}
\usepackage{graphicx}

\usepackage{array} 

\definecolor{orange}{rgb}{1,0.5,0}

\begin{document}
	
\title{\LARGE \bf
TacEva: A Performance Evaluation Framework For Vision-Based Tactile Sensors
}

\author{
Qingzheng Cong†, Steven Oh†, Wen Fan†, Shan Luo, Kaspar Althoefer, Dandan Zhang*
\thanks{
D. Zhang, W. Fan, and Q. Cong are with the Department of Bioengineering, Imperial College London, UK (emails: \{d.zhang17, w.fan24, qcong\}@imperial.ac.uk). 
S. Oh is with Waseda University, Japan (email: oh.steven@fuji.waseda.jp). 
S. Luo is with the Department of Engineering, King’s College London, UK (email: shan.luo@kcl.ac.uk). 
K. Althoefer is with the Centre for Advanced Robotics, Queen Mary University of London, UK (email: k.althoefer@qmul.ac.uk). 
† These authors contributed equally to this work. 
* Corresponding author: D. Zhang (email: d.zhang17@imperial.ac.uk).
}
}

\maketitle

\begin{abstract}
Vision-Based Tactile Sensors (VBTSs) are widely used in robotic tasks because of the high spatial resolution they offer and their relatively low manufacturing costs. However, variations in their sensing mechanisms, structural dimension, and other parameters lead to significant performance disparities between existing VBTSs. This makes it challenging to optimize them for specific tasks, as both the initial choice and subsequent fine-tuning are hindered by the lack of standardized metrics. To address this issue, TacEva is introduced as a comprehensive evaluation framework for the quantitative analysis of VBTS performance. The framework defines a set of performance metrics that capture key characteristics in typical application scenarios. For each metric, a structured experimental pipeline is designed to ensure consistent and repeatable quantification. The framework is applied to multiple VBTSs with distinct sensing mechanisms, and the results demonstrate its ability to provide a thorough evaluation of each design and quantitative indicators for each performance dimension. This enables researchers to pre-select the most appropriate VBTS on a task by task basis, while also offering performance-guided insights into the optimization of VBTS design. A list of existing VBTS evaluation methods and additional evaluations can be found on our website:
\url{https://stevenoh2003.github.io/TacEva/}.
\end{abstract}

\section{Introduction}

Robots have yet to attain the level of manipulative dexterity exhibited by humans, a challenge rooted in the difficulty of accurately acquiring detailed contact information in physical environments \cite{billard2019trends}. Tactile sensing has therefore become indispensable for delicate and precise robotic manipulation in embodied intelligence systems \cite{F-TAC}. A notable development in this domain has been the rise of vision-based tactile sensors (VBTSs) \cite{yuan2017gelsight}. These sensors employ high-resolution cameras to capture detailed contact surface information, thereby integrating seamlessly with computer vision and image-based deep learning methods. Nevertheless, among VBTSs, we see a wide variety of architectures, structural dimensions, and fabrication techniques, depending on the exact nature of the application requirements  \cite{zhang2022hardware}.

The rapid development of VBTSs has created an urgent need for standardized performance evaluation. Selecting an appropriate VBTS for a specific task scenario remains challenging, as distinct sensor designs offer substantial variation in performance. A universal evaluation protocol would therefore facilitate fair comparison across sensor designs, support informed selection, and guide design optimization. However, a significant gap persists in the field: no standardized framework currently exists for VBTS evaluation, and the inconsistency of current metrics limits objective, reproducible, and comprehensive cross-sensor assessment. This challenge is further compounded by the inherently multi-modal nature of tactile sensing \cite{zhang2025design}, which necessitates coordinated evaluation across multiple performance dimensions.


\begin{figure}
    \centering
    \includegraphics[width=1.0\linewidth]{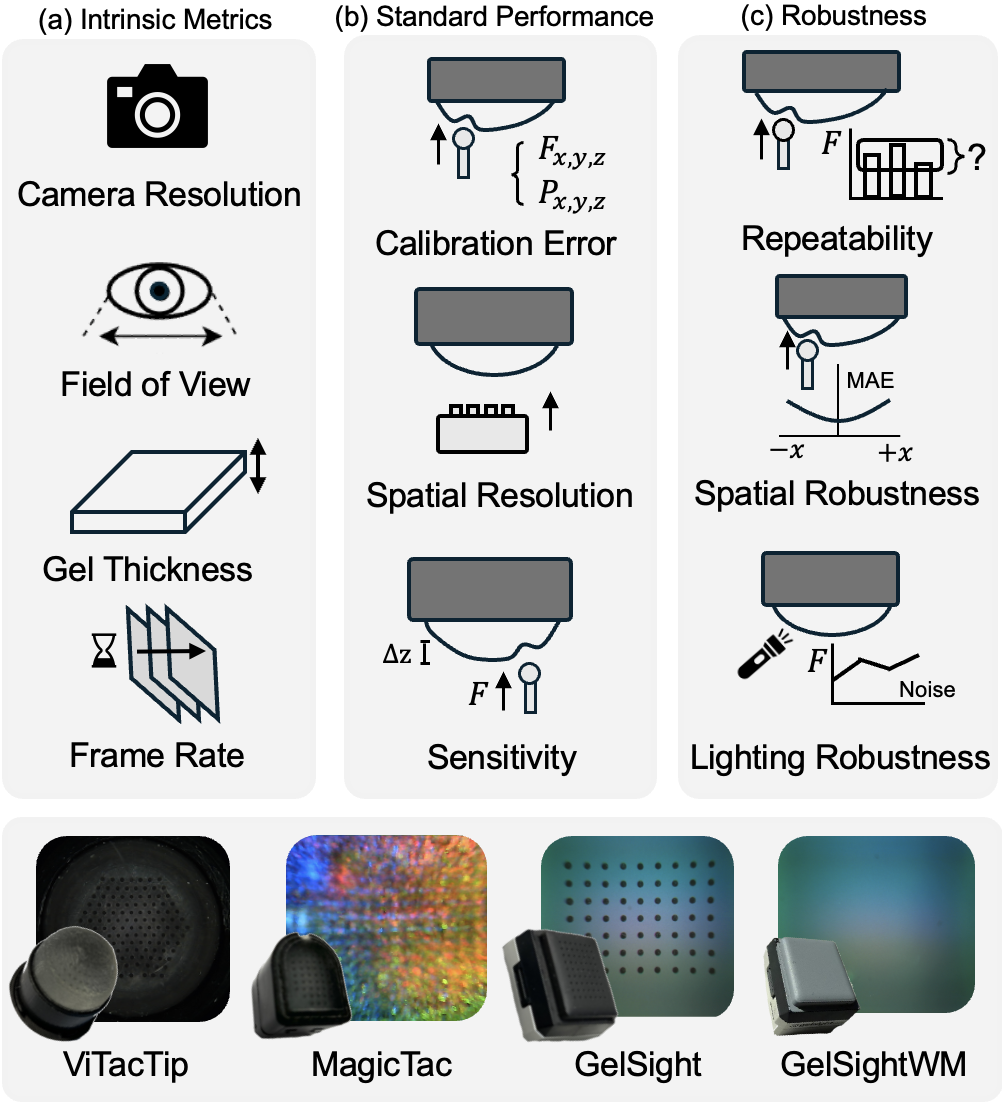}
    \captionsetup{font=footnotesize}
\caption{\textbf{TacEva benchmarking pipeline.} A unified framework for evaluating VBTSs that enables systematic comparison across (a) intrinsic metrics, (b) standard performance, and (c) robustness. We demonstrate TacEva on four representative VBTSs: ViTacTip\cite{fan2024vitactip}, MagicTac\cite{fan2024magictac}, GelSight Mini Tracking Marker Gel (GelSight), and GelSight Mini Standard Silicone Gel (GelSightWM)\cite{yuan2017gelsight}}
    \label{fig:teaser}
        \vspace{-0.5cm}
\end{figure}

The application scenarios for VBTSs are inherently diverse. This makes it difficult to establish universal, broadly applicable evaluation metrics that remain meaningful across the field. Further complexity arises from the fact that VBTSs are commonly fabricated using silicone elastomers, introducing additional considerations related to soft-material mechanics, optics, and imaging. Owing to differences in their underlying sensing mechanisms, each VBTS design inevitably exhibits its own profile of strengths and weaknesses. As a result, certain positive performance characteristics will typically tend to be emphasized, while weaker ones may be under-reported. This is likely to complicate informed selection, making it more challenging for prospective users. 

To address these challenges, we propose TacEva, a systematic evaluation framework that integrates performance quantification with a structured and reproducible assessment pipeline. TacEva is designed to provide consistency and comparability across VBTS designs, enabling practitioners to make evidence-based decisions while obtaining a holistic understanding of sensor performance (see Fig. \ref{fig:teaser}). It also offers sensor developers clear guidance for targeted optimization during the design process. By defining and standardizing a comprehensive set of performance metrics and evaluation protocols, TacEva aims to provide a unified and transparent characterization of VBTS designs, thereby facilitating objective comparison, reliable selection, and more informed innovation for future sensor development.

\section{Related Work}

\subsection{Development of Vision-Based Tactile Sensors}

Several works have reviewed and summarized the development of VBTSs from the perspectives of structural design \cite{zhang2022hardware} and sensing mechanisms \cite{he2022review}, and have also demonstrated their usability in diverse robotic applications \cite{fang2024force}.
Building on these reviews, the CrystalTac family of sensors \cite{fan2024crystaltac} offers a systematic classification of VBTSs according to their sensing principles. These include: (i) Intensity Mapping Method (IMM), which infers contact geometry and pressure through spatial variations in reflected light intensity; (ii) Marker Displacement Method (MDM), which detects surface deformation by tracking the displacement of embedded markers under force; and (iii) Modality Fusion Method (MFM), which employs transparent skin to enable multi-modal perception.

GelSight-type sensors \cite{yuan2017gelsight}, for example, are IMM-based and are widely used for reconstructing fine contact geometry, enabled by a reflective coating on a compliant surface. Later updates also incorporate markers on the gel surface, thereby combining both IMM with MDM to improve deformation tracking and force regression. TacTip-type sensors are MDM-based and capture dynamic features such as force, torque, and vibration by tracking markers with geometric patterns \cite{ward2018tactip}. Recent work increasingly explores multi-modal extensions, e.g., MFM-based sensors, such as those that combine vision with tactile sensing \cite{fan2024vitactip}, to enhance versatility. Another example is the F-TOUCH sensor \cite{FTOUCH} that unifies six-axis force/torque sensing and tactile sensing within a single vision-based device. However, as VBTS designs become more complex, so does their evaluation, underscoring the need for standardized assessment frameworks.


\subsection{Evaluation of tactile sensors}

\subsubsection{Non-vision-based tactile sensors}
Non-vision-based tactile sensors, such as magnetic (uSkin, Hall-effect arrays) \cite{uskinwaseda,magnetic_2016}, capacitive (iCub fingertips \cite{schmitz2010tactile}, PPS TactArray \cite{trejos2009robot}), piezoresistive (Weiss WTS arrays \cite{stassi2014flexible}), piezoelectric (PVDF-based skins \cite{pei20233d}), impedance/fluidic (BioTac \cite{fishel2012sensing}), and barometric/fluidic (BaroTac \cite{kim2022barotac}), all have established, well-defined evaluation practices. The following examples illustrate this. uSkin is evaluated in terms of its pre-calibration response, including axis crosstalk, overload durability, calibration accuracy (via $R^2$ comparisons across fitting models), hysteresis during load–unload cycles, and signal-to-noise ratio (SNR) \cite{uskinwaseda}. Capacitive fingertips are typically assessed by pre-/post-molding SNR, accumulated-load nonlinearity and hysteresis, temperature-drift compensation, and full triaxial calibration \cite{tricapacitive}. Piezoelectric skins report measurable force range, axial sensitivity, nonlinearity, repeatability, frequency response, and mechanical flexibility \cite{yu2016flexible,pei20233d}. Recent custom designs also quantify force-prediction accuracy and robustness to magnetic noise \cite{pattabiraman2025}. 
As can be seen, metrics to evaluate tactile sensors typically include force/torque range and resolution, spatial density, bandwidth, slip/texture sensing, durability, and calibration fidelity, and, in some cases, certain more bespoke quantifiers. That said, non-vision-based tactile sensing does, overall, have relatively standardized protocols.

\subsubsection{Vision based tactile sensors}

Comprehensive frameworks that integrate both design-level and task-level evaluations are rare in the VBTS domain. Existing studies of VBTSs typically follow one of two approaches - hardware-focused or task-focused. Hardware-focused studies compare design properties such as sensing area, dynamic range, footprint, mass, and other parameters, across new prototypes \cite{taylor2022gelslim,geltip,song2024satac}. Task-focused studies judge performance via application-specific metrics such as object-classification accuracy \cite{fan2022graph}, contact-shape reconstruction error \cite{do2022densetact}, force/pose regression error \cite{fan2023tac}, slip-detection or manipulation of objects success rate \cite{Tacman}. Early sensor-agnostic protocols for force resolution and accuracy \cite{vbtseval2002} pre-date today’s data-driven methods, while more recent efforts refine  design-related criteria \cite{Agarwal_2025} or focus solely on task metrics \cite{sparsh2024}. To bridge this gap, we introduce a unified framework that integrates design properties with indicators that best predict real-world task performance.


\section{TacEva Framework Overview}
\label{sec:taceva}
The pipeline evaluates VBTS from three primary aspects: intrinsic metrics (Fig. \ref{fig:teaser}(a)), standard performance (Fig. \ref{fig:teaser}(b)), and robustness (Fig. \ref{fig:teaser}(c)). Sec. \ref{sec:taceva-intrinsic} formalizes the intrinsic properties. Sec. ~\ref{sec:force_calib}, \ref{sec:taceva-spatial}, and \ref{sec:taceva-sensitivity}
detail the experimental protocol and metrics for standard performance. Lastly, Sec. \ref{sec:taceva-spatial-robustness}, \ref{sec:taceva-lighting-robustness}. \ref{sec:taceva-repeatability} define the robustness criteria and describe the corresponding experimental procedures.

\subsection{
Intrinsic Properties}
\label{sec:taceva-intrinsic}

\subsubsection{Definitions}
Properties intrinsic to a VBTS's hardware such as camera and elastomer are classified as intrinsic metrics. These properties of a VBTS can be categorized as follows:
\begin{itemize}
    \item \textbf{Camera resolution:} The pixel resolution of the internal camera. Higher resolution captures finer surface deformations, improving contact localization and shape reconstruction.
    \item \textbf{Gel thickness:} The thickness (in mm) of the compliant elastomer layer, typically measured at the dome’s center. A thicker gel increases compliance and durability but may reduce spatial resolution and introduce optical distortion.
    \item \textbf{Field of view (FOV):} The effective sensing area (in $\mathrm{mm}^2$) of the tactile surface that is visible to the internal camera. A larger FOV allows coverage of a greater portion of the elastomer surface, improving the detection of distributed or large-area contact patterns. In this work, we report projected area rather than angular FOV, as it directly reflects the usable sensing region and enables comparison across different sensor geometries.

    \item \textbf{Frame rate (Frames Per Second)}: The image-capture frequency (in Hz) of the internal camera. Higher FPS improves temporal resolution to track fast or dynamic contact events.
\end{itemize}

These properties describe the intrinsic hardware characteristics of a VBTS, which set the fundamental limits on sensing quality and form the basis for evaluating higher-level performance.


\subsection{Calibration} \label{sec:force_calib}
\begin{figure*}[t]
    \centering
    \includegraphics[width=1\linewidth]{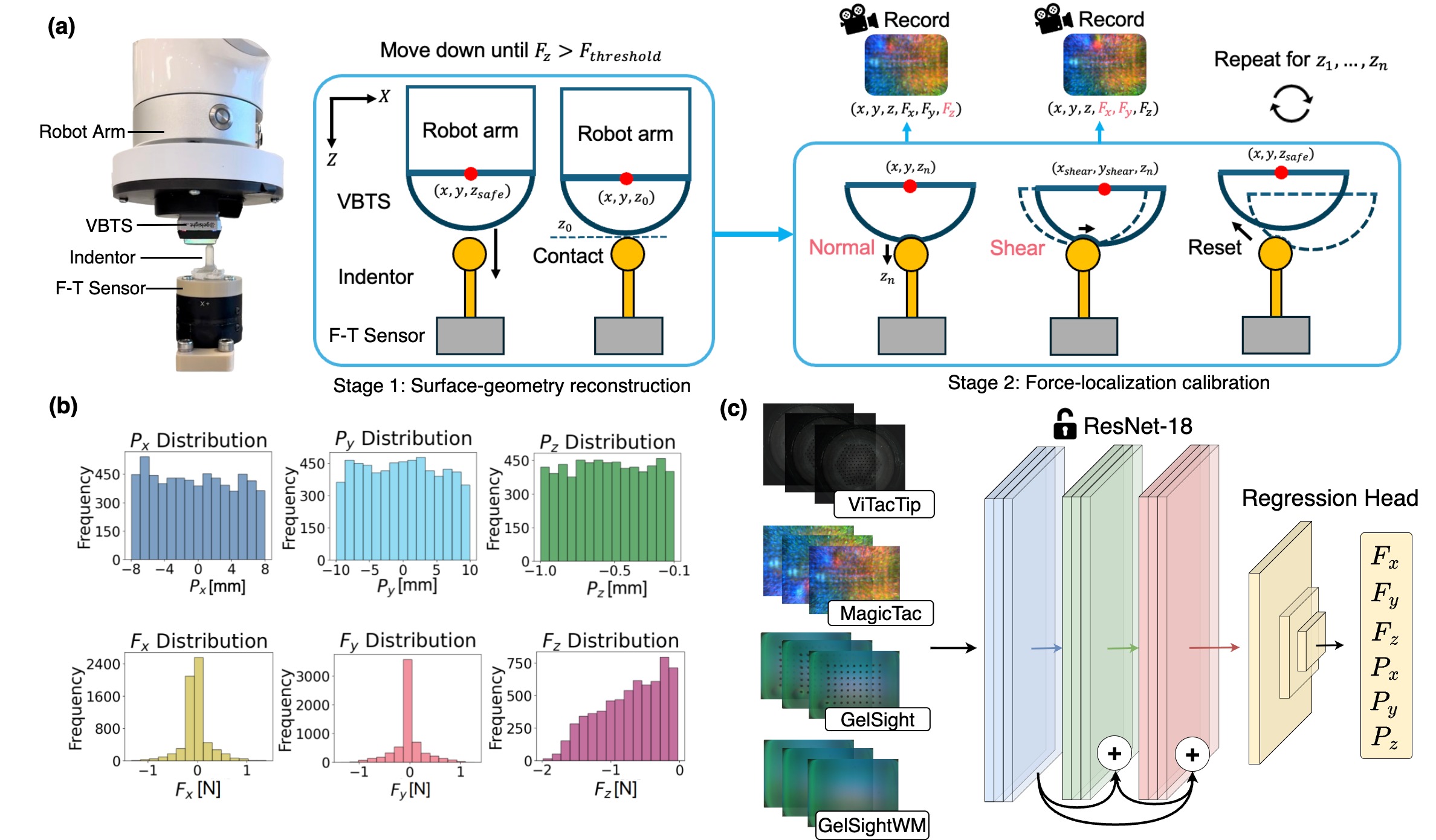}
        \captionsetup{font=footnotesize}
    \caption{\textbf{Calibration procedure.} (a) Experimental data collection. Tool positions $(P_x,P_y)$ were uniformly sampled across the sensor surface at a fixed insertion depth $P_z$, then randomly perturbed along the $x$, $y$, and $z$ axes to generate the dataset. Our framework implements a two-stage calibration procedure: \textbf{Stage 1.} Surface-geometry reconstruction to map the 3D sensor surface using a robotic arm and force-sensitive spherical indenter (10mm radius), and \textbf{Stage 2.} Force-localization calibration establishes optical-output-to-force mappings through systematic probing. The two stages are done sequentially over a large set of $(x, y)$ points across the sensor surface, simultaneously providing both surface reconstruction results and data for optical-output-to-force mapping. (b) The histograms show empirical distributions of the three position components ($P_x$, $P_y$, $P_z$) along with the corresponding force measurements ($F_x$, $F_y$, $F_z$). (c) Regression pipeline. A ResNet-18 backbone (trained from scratch with all layers unfrozen) extracts features from the tactile images, and a lightweight regression head simultaneously predicts the six-dimensional output vector $(P_x,P_y,P_z,F_x,F_y,F_z)$. }
        \label{fig:calibration_procedure}
    \vspace{-0.5cm}
\end{figure*}
\subsubsection{Definitions}
Accurate characterization of a VBTS requires calibrating both its surface geometry and its mapping from raw optical outputs to contact forces and positions. 
This is achieved through two sequential procedures with the sensor mounted on a robotic manipulator: (1) \emph{surface-geometry reconstruction}, which recovers the three-dimensional shape of the sensor surface to localize contact points precisely; and (2) \emph{force-localization calibration}, which maps the raw optical outputs of the sensor to ground-truth forces and indentation positions.

\subsubsection{Experimental Protocols}
 To obtain these calibration results, a six-axis force/torque transducer (M3813B) is rigidly affixed to the workbench, and a 2-mm-radius, 3D-printed spherical indenter is attached to its sensing head. The VBTS is then mounted on a robot arm for data collection.

As shown in Fig. \ref{fig:calibration_procedure} (Stage 1), the robot arm moves the indenter toward the sensor surface, along the vertical axis (z-axis), in small steps of 0.1 mm until contact is detected. The corresponding end-effector position at the moment of initial contact is recorded as a point on the sensor surface. 
 In Stage 2 of Fig.~\ref{fig:calibration_procedure}, after each initial contact point detected in Stage~1, the robot performs four additional indentations to randomized depths $z_n$ in the vertical direction. At each indentation depth, a small displacement is applied in the $x\!-\!y$ plane to introduce variation in the contact direction and to capture both normal and shear force responses. Note that the maximum indentation depth and normal force were constrained by the safe operating range of each sensor to avoid destructive loading. 
For example, the ViTacTip was limited to $\sim$3.5\,mm ($F_z \leq 0.7$\,N), MagicTac to $\sim$1.2\,mm ($F_z \leq 2.85$\,N), and GelSight-series to $\sim$1.0\,mm ($F_z \leq 1.83$\,N). 
These limits were established empirically from preliminary trials and consistently adopted in all calibration and evaluation experiments.

In our work, the contact positions based on the robot kinematics \((P_x, P_y, P_z)\) and corresponding six-axis force measurements from the F/T sensor \((F_x, F_y, F_z\)) serve as ground-truth labels. Each probe instance is paired with a synchronized raw camera image from the VBTS, creating a labeled dataset for supervised learning.

We aggregate exhaustive and randomized samples across the entire sensing area to form a comprehensive dataset. Before training, all labels \((F_x,F_y,F_z,P_x,P_y,P_z)\) are min–max normalized to equalize their dynamic ranges. The dataset is partitioned into training, validation, and test sets in a 70:20:10 split. We employ a ResNet-18 backbone for the regression backbone. The network is trained from scratch with randomly initialized weights and optimized with the SGD optimizer for 100 epochs. Model selection is based on the lowest validation error, and final evaluation compares performance metrics across different VBTSs. The resulting ResNet-18 regression model trained from scratch serves as a baseline to compare sensor performance.

Lastly, we evaluate the predictive accuracy for force and contact position across different sensors using the following metrics:

\noindent \textbf{Mean Absolute Error} (MAE):  
Measures the average absolute difference between predictions and ground truth.
\begin{equation}
\mathrm{MAE} = \frac{1}{n} \sum_{i=1}^{n} \lvert y_i - \hat{y}_i \rvert ,
\end{equation}

\noindent \textbf{Coefficient of Determination} (\(R^2\)):  
Indicates how well predictions align with actual data trends.
\begin{equation}
\mathrm{R}^2 = 1 - \frac{\sum_{i=1}^{n} (y_i - \hat{y}_i)^2}{\sum_{i=1}^{n} (y_i - \bar{y})^2} ,
\end{equation}

\noindent \textbf{Symmetric Mean Absolute Percentage Error}  (sMAPE): 
Quantifies relative error as a percentage by normalizing absolute prediction errors against the average of the true and predicted values:
\begin{equation}
\mathrm{sMAPE} = \frac{1}{n} \sum_{i=1}^{n} \frac{\lvert y_i - \hat{y}_i \rvert}{\tfrac{\lvert y_i \rvert + \lvert \hat{y}_i \rvert}{2} + \epsilon} \times 100\% .
\end{equation}

\medskip
\noindent Here, \(y_i\) denotes the ground-truth value, \(\hat{y}_i\) the predicted value, \(n\) the total number of samples, 
\(\bar{y} = \tfrac{1}{n}\sum_{i=1}^{n} y_i\) the mean of the ground-truth values, and \(\epsilon = 1 \times 10^{-8}\) a small constant to prevent division by zero.

MAE provides an intuitive measurement of average absolute deviations between predictions and ground truth, directly indicating the magnitude of prediction errors. $R^2$ complements MAE by quantifying the proportion of variance explained by the model. Both MAE and $R^2$ have been widely used in tactile sensing regression tasks \cite{fan2024vitactip}.

sMAPE highlights relative error as a percentage, normalizing prediction errors against signal magnitude. This makes sMAPE particularly effective at highlighting proportional errors and comparing accuracy across different scales. Although less commonly reported in VBTS literature, we introduce it here to complement MAE and $R^2$ for cross-sensor evaluation.

\subsection{Spatial Resolution (\texorpdfstring{$SR(\varepsilon)$}{SR(epsilon)})}
\label{sec:taceva-spatial}
\subsubsection{Definitions}
 The spatial resolution of a VBTS denotes the minimum distance between two points on the sensor surface that can be distinctly sensed as separate touch points. This reflects the sensor's capacity to differentiate between closely spaced stimuli. A higher spatial resolution score means the sensor can detect finer details of the touched object or surface, making this aspect critical for applications that require detailed tactile feedback, such as dexterous robotic manipulation or surface texture recognition.

\begin{figure*}[t]
    \centering
    \includegraphics[width=1.0\linewidth]
    {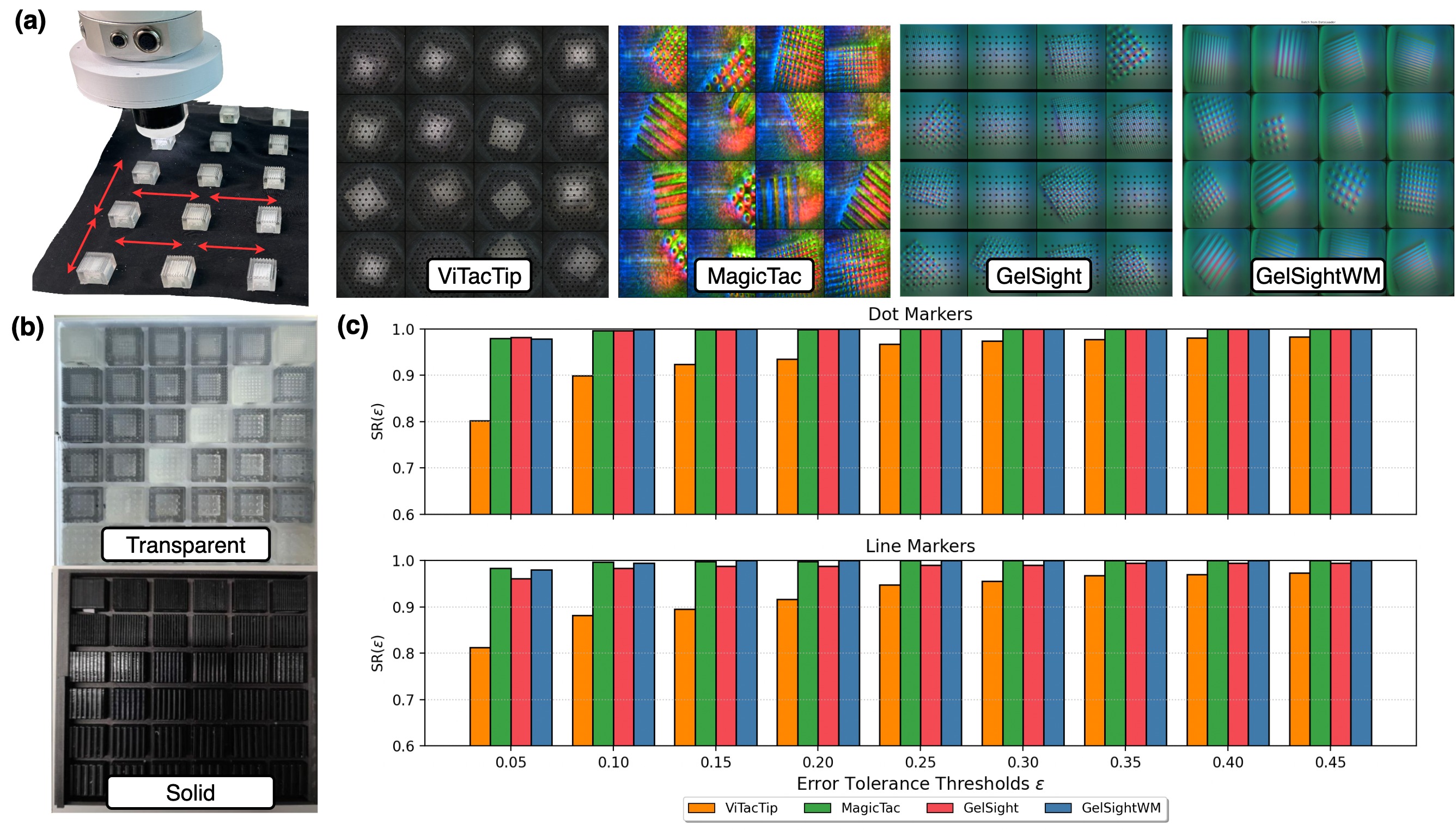}
     \captionsetup{font=footnotesize}
    \caption{(a) Data collection setup using grating boards and raw visual output from selected VBTSs used in the experiment. A black cloth base and sufficient spacing between the grating boards mitigated lighting or positional bias in data collection and prevented transparent VBTSs from exploiting background patterns as cues. (b) A total of 72 3D printed line and dot grating boards of both transparent and solid materials were used to conduct the spatial resolution evaluation. 
    (c) Classification results of VBTSs showing the effect of different error tolerance thresholds on prediction accuracy (spatial resolution results at higher error tolerance thresholds are omitted, as they led to near-perfect classification performance.)} 
    \label{fig:spatial-resolution}
  \vspace{-0.5cm}
\end{figure*}

\subsubsection{Experimental Protocols}
We used an Objet500 3D printer to fabricate test samples with linear and cylindrical grating features (in Fig. \ref{fig:spatial-resolution}(b)), with grating widths and spacings ranging from 0.25 mm to 1.75 mm in 0.05 mm increments. The VBTS was mounted onto the robot’s end-effector and used to press each sample 100 times while varying the tool center point (TCP) yaw orientation, producing data as shown in Fig. \ref{fig:spatial-resolution}(a) right. The entire process is automated with a large number of grating samples placed onto a 3D printed board with equal spacing (in Fig. \ref{fig:spatial-resolution}(a), the red arrow). This protocol yielded a diverse dataset capturing the sensor's response to samples with different spatial features and orientations.

We used the dataset to train a classifier to distinguish between samples with different grating resolutions. After training, we evaluated accuracy under varying error tolerance thresholds \(\varepsilon\), where \(\varepsilon\) specifies the maximum allowable deviation (in mm) between the predicted and ground-truth resolution levels. Since the gratings' dimensions were fabricated in 0.05\,mm increments, all thresholds were chosen as integer multiples of 0.05\,mm, ranging from the strictest level (0.05\,mm) to more relaxed values (e.g., 0.50–1.00\,mm).
 Rather than reporting a single scalar metric such as cross-entropy loss, we define Spatial Resolution $SR(\varepsilon)$ as the curve of correctly classified samples accuracy over thresholds \(\varepsilon\):
\begin{equation}
SR(\varepsilon) 
= \frac{1}{N} \sum_{i=1}^{N} 
\begin{cases}
1, & \text{if } \lvert \hat{r}_i - r_i \rvert \leq \varepsilon, \\
0, & \text{otherwise}.
\end{cases}
\label{eq:resolution}
\end{equation}
where \(\hat{r}_i\) and \(r_i\) denote the predicted and ground-truth resolution levels of the \(i\)-th sample.

In this way, $SR(\varepsilon)$ provides a comprehensive characterization, revealing how performance degrades as tolerance becomes stricter (i.e., as \(\varepsilon \downarrow\)). In addition, it directly shows how VBTS performance varies under different resolution requirements, offering a more intuitive and physically meaningful comparison. 
This is particularly useful for practical tasks, since not all applications demand the finest resolution, and sensors can be compared under tolerance levels aligned with task needs. 

\subsection{Mechanical Sensitivity (\texorpdfstring{$S$}{S})}
\label{sec:taceva-sensitivity}

\subsubsection{Definitions}
\emph{Mechanical sensitivity} refers to the compliance of the sensor surface under normal loading, effectively quantifying the indentation response per unit applied force. We define the mechanical sensitivity \(S\) as the ratio of indentation depth $\Delta z$ along the surface normal to the corresponding applied normal force $F$:
\begin{equation}
S = \frac{\Delta z}{F}. 
\end{equation}
This definition enables the generation of sensitivity maps across the sensor surface to assess local responsiveness under normal loading. This metric is chosen because it provides a physically interpretable measure of compliance, directly relating normal indentation to applied force. This concept has been highlighted in previous tactile sensor studies in TacTip and GelSight \cite{yuan2017gelsight, donlon2018gelslim,ward2018tactip}, where the compliance of the sensor was shown to strongly affect sensing performance, such as force response, durability, and the fidelity of contact features. 

Additionally, to quantify the \emph{uniformity} of the sensitivity distribution, we divide the $(x,y)$ plane into uniform bins and compute the mean $\mu$ and standard deviation $\sigma$ of the average sensitivity values across bins. The uniformity score is then defined as
\begin{equation}
U = \frac{1}{1 + \sigma / |\mu|} \;\in (0,1],
\end{equation}
where a larger $U$ indicates greater uniformity.
This formulation is mathematically equivalent to a normalized inverse of the coefficient of variation, a standard statistical measure of relative dispersion. 
Uniform sensitivity distribution is fundamental for reliable VBTS performance, because inconsistent sensitivity can lead to location-dependent bias, distort force estimation, and ultimately reduce the robustness and reliability of tactile perception in tasks like force regression or contact reconstruction \cite{yuan2017gelsight,lee2024behavioral}.

\subsubsection{Experimental Protocols}
Sensitivity is derived from the calibration dataset (defined in Sec. \ref{sec:force_calib}). Each calibration trial provides a contact position \((P_x,P_y,P_z)\) along with ground-truth forces \((F_x,F_y,F_z)\) measured by a six-axis force/torque (F/T) sensor. 
All probe samples across the sensing surface are aggregated and projected onto the \((x,y)\) plane. The plane is partitioned into spatial bins, and for each bin we compute the mean sensitivity, which can be visualized as a heatmap. This procedure yields a spatial sensitivity map of the entire sensor surface under normal loading.

\subsection{Spatial Robustness (\texorpdfstring{$R_{\text{spatial}}$}{Rspatial})}

\label{sec:taceva-spatial-robustness}

\subsubsection{Definitions}
Spatial robustness refers to how consistently the model performs when the point of contact moves to different locations on the sensor surface. A model with high spatial robustness will give nearly identical outputs regardless of whether the contact is near the center, the edge, or any intermediate point. This is crucial to ensure that the system behaves predictably in real-world scenarios in which the exact contact position cannot be controlled.
For each channel $c$, the sensor surface (radial distance) and indentation range are divided into uniform bins. In practice, both radial distance and depth were uniformly divided into bins with a step size of 0.01 in the normalized range (approximately 100 bins over [0,1]).
Within each bin, the MAE is computed, and the standard deviation (STD) of these bin means is used as the variability measure. The STD values were then smoothed and averaged to obtain a single robustness score for each channel. 
This is defined as the combined average of distance and depth variability:
\begin{equation}
R_{\text{spatial},c} \;=\; \tfrac{1}{2}\Big(
\mathrm{STD}\big(\{m^{\text{dist}}_b\}\big) \;+\;
\mathrm{STD}\big(\{m^{\text{depth}}_d\}\big)
\Big),
\end{equation}
where $m^{\text{dist}}_b$ and $m^{\text{depth}}_d$ are the mean errors in each distance and depth bin, respectively.
Smaller $R_{\text{spatial},c}$ indicates higher robustness.
\begin{figure*}
    \centering
    \includegraphics[width=1\linewidth,height=7.5cm]{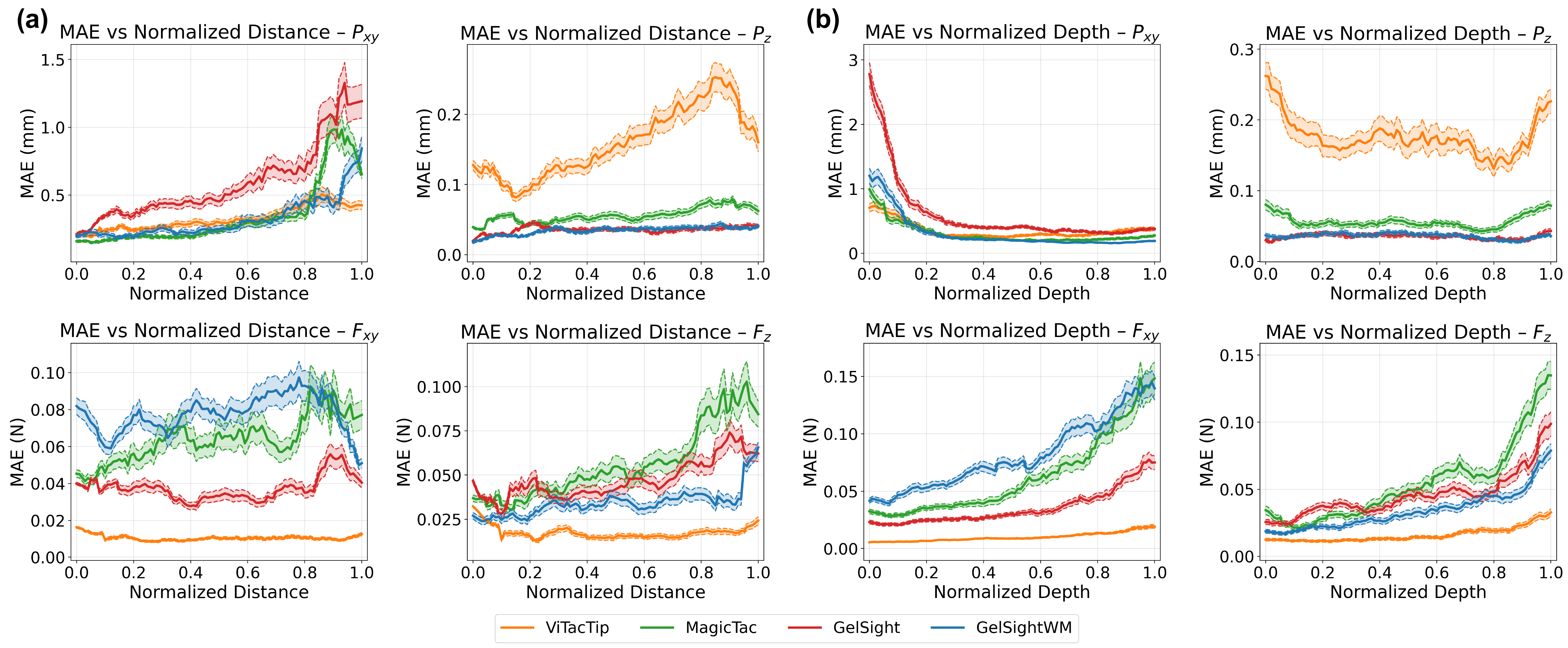}
      \captionsetup{font=footnotesize}
    \caption{Smoothed MAE distribution of all parameters for four sensors as a function of spatial coordinates: (a) MAE distribution over normalized distance from the center point; (b) MAE distribution over normalized depth. Solid lines show the rolling‐mean MAE, and shaded bands are ± 0.1 × STD. These distributions form the basis for computing the spatial robustness score $R_{\text{spatial},c}$, defined as the average STD of mean within-bin errors along distance and depth. As all x-axis values are normalized, the original physical ranges are reported here for reference: Gelsight (distance: 12.5 mm; depth: 1.0 mm); GelsightWM (distance: 12.5 mm; depth: 1.0 mm); ViTacTip (distance: 17 mm; depth: 3.5 mm); MagicTac (distance: 10 mm; depth: 1.0 mm).}
    \vspace{-0.5cm}
    \label{fig:Spatial robustness}
\end{figure*}

\subsubsection{Experimental Protocol}
We collected a new dataset for each sensor comprising 1,600 indentation points, following the same calibration protocol as in Section~\ref{sec:force_calib}. Given the data in our study, model predictions were compared against ground truth to compute MAE and variability (STD). 

These errors were then post-processed to examine their distribution along two dimensions: (i) radial distance from the sensor center, reflecting surface uniformity (Fig.~\ref{fig:Spatial robustness}(a)); and (ii) normalized indentation depth (Fig.~\ref{fig:Spatial robustness}(b)), reflecting depth uniformity. The resulting trends provide a direct measure of how consistently each sensor maintains accuracy across its spatial domain.
\subsection{Lighting Robustness (\texorpdfstring{$R_{\text{light}}$}{Rlight})}
\label{sec:taceva-lighting-robustness}
\subsubsection{Definitions}

Lighting robustness refers to a model’s ability to maintain stable performance under varying illumination conditions, such as bright daylight, dim indoor lighting, or dynamically changing environments. This metric is only applicable to transparent or semi-transparent VBTS devices; opaque sensors inherently block external light and are therefore not influenced by lighting variations (their robustness score is nominally deemed as 1, but they are effectively excluded from this evaluation).

We quantify robustness by measuring the model’s deviation in prediction accuracy across different lighting conditions and comparing these to a baseline (the original training setting). The metric is defined as follows:

\begin{equation}
\text{R}_{light} = \frac{\left|\dfrac{I_{c}}{I_{o}} - 1\right|}
{\left|\dfrac{I_{c}}{I_{o}} - 1\right| + \left|\dfrac{\text{MAE}_{c}}{\text{MAE}_{o}} - 1\right|}
\end{equation}
where \(\mathrm{MAE}_\text{o}\) is the Mean Absolute Error under the original (training) lighting condition; \(\mathrm{MAE}_\text{c}\) is the Mean Absolute Error under test lighting condition \(c\); \(I_\text{o}\) is the mean light intensity in the original lighting condition; \(I_\text{c}\) is the mean light intensity in the test lighting condition \(c\).

A smaller relative change in prediction error as compared to the normalized lighting difference 
\(\left(\tfrac{I_c}{I_o}-1\right)\)indicates minimal performance degradation relative to changes in lighting intensity. Consequently, a higher robustness score \(R_c\) (approaching 1) reflects stronger resistance to lighting variations.

\subsubsection{Experimental Protocol}

To assess the robustness of the lighting, we introduced controlled variations in external illumination during the testing, distinct from the training illumination condition used in calibration in section \ref{sec:force_calib}. These included diffuse overhead lighting and point-source illumination, in different combinations and at varying intensities, resulting in four distinct ambient lighting setups (denoted as scenes S1-S4). Specifically, \textbf{S1} denotes a diffuse ceiling source that produces uniform overhead illumination; \textbf{S2} employs a lateral wide-beam point source creating broad angular coverage; \textbf{S3} utilizes a ring-shaped emitter mounted beneath the sensor to provide circumferential lighting around the camera; and \textbf{S4} applies a narrow-beam directional point source that produces a concentrated spot of illumination. We characterized each lighting condition by the \textbf{mean gray-scale intensity} of the captured images, providing a relative proxy for illumination level across sensors. The baseline model (ResNet-18) trained during calibration was then directly applied to these test sets collected under varied lighting conditions, and its predictive performance under each condition was used to compute the corresponding robustness metrics.

\subsection{Repeatability (\texorpdfstring{${Rep}$}{Rep})}
\label{sec:taceva-repeatability}

\subsubsection{Definitions}
Repeatability refers to the ability of the sensor to give consistent readings under the same conditions over multiple occasions (i.e., when the same stimulus is applied multiple times).
To quantify repeatability, we compute the standard deviation of $N$ repeated 
measurements at each spatial test point and indentation depth, and then average across all test conditions. Specifically, for each output channel $c$ (either positional $P_x,P_y,P_z$ or force $F_x,F_y,F_z$), the standard deviation across $N$ repeated trials is first computed at every spatial point and indentation depth.
For each spatial test point $k = 1,\ldots,K$ and indentation step $d = 1,\ldots,D$, $N$ repeated trials are performed, yielding outputs $\hat{c}_{k,d,r}$ ($r=1,\ldots,N$). 
The repeatability score is then defined as
\begin{equation}
\mathrm{Rep}_c \;=\; \frac{1}{KD}\sum_{k=1}^{K}\sum_{d=1}^{D} 
\mathrm{STD}\!\left(\hat{c}_{k,d,1},\ldots,\hat{c}_{k,d,N}\right).
\end{equation}
A smaller $\mathrm{Rep}_c$ indicates higher repeatability. 

\subsubsection{Experimental Protocol}
A total of $K=100$ random surface points were selected on the sensor surface, and each point was sampled with 10 repeated trials (N = 10). At each point, the sensor was gradually pressed down in steps of 0.1 mm until reaching the predefined maximum depth used during model training, yielding $D$ indentation steps. 
For every $(k,d)$ combination, $N=10$ repeated trials were conducted, with the sensor pressed to the target depth and fully retracted before the next trial. This procedure provides both a global repeatability score $\mathrm{Rep}_c$ and depth-dependent STD curves for visualization.

\section{Case Studies and Evaluation}
In this section, we evaluate four representative VBTSs using the evaluation pipeline defined in Sec.~\ref{sec:taceva}, present the results, and provide a comparative analysis of their performance.

\subsection{Justification of Sensor Selection}

To ensure comprehensive coverage of sensing modalities, we selected four representative VBTSs, as illustrated in  Fig.\ref{fig:teaser} (top):
\begin{itemize}
\item \textbf{ViTacTip}: A transparent sensor with embedded markers integrating MDM and MFM \cite{fan2024vitactip}. It introduces multi-modal perception through its transparent elastomer. This not only enhances deformation estimation but also enables proximity sensing, making it a suitable candidate for evaluating modality fusion.
\item \textbf{MagicTac}: a recent design that integrates IMM, MDM, and MFM simultaneously by using a 3D multi-layer grid-based elastomer \cite{fan2024magictac}. This design combines the photometric cues from internal light reflections (IMM), the partial transparency that enables proximity and multi-modal perception (MFM), and the continuous markers provided by the embedded grid (MDM). 
\item \textbf{GelSight (standard)}: An extension of GelSightWM, GelSight incorporates markers on the elastomer surface, thereby combining IMM and MDM. This enables comparison between marker-less and marker-based implementations, highlighting the impact of marker tracking on deformation sensing.
\item \textbf{GelSightWM}: a marker-less, opaque design that is based purely on IMM \cite{yuan2017gelsight}, representing the most widely adopted class of VBTSs in which dense surface textures are captured through light-intensity variations. This sensor provides a baseline for assessing the performance of marker-less, intensity-only sensing (IMM).

\end{itemize}
These sensors span the core design spectrum of VBTSs and provide a diverse testbed for validating the generality and applicability of the proposed TacEva framework. For benchmarking, we employ ResNet-18 as a common backbone across all four sensors for the calibration process (in Fig. \ref{fig:calibration_procedure}). Although this offers a consistent baseline, it does not represent state-of-the-art performance and we would therefore encourage the customization of models tailored to specific sensor characteristics to fully exploit the framework.

\subsection{Experimental Setup}

By illustrating TacEva courtesy of case studies using the four VBTSs listed above (which collectively cover the primary sensing modalities), we demonstrate that the framework can accommodate diverse designs and enable meaningful, like-for-like comparison between sensors. All experiments were conducted with a UFactory~850 robot arm, and model training was performed on an NVIDIA RTX~4080 GPU. The intrinsic metrics (physical and optical specifications)  of the VBTS are summarized in Table~\ref{tab:vbts_specs}.

\begin{table}[ht]
  \centering
  \caption{Intrinsic metrics of representative VBTSs}
  \label{tab:vbts_specs}
  \small
  \setlength{\tabcolsep}{6pt} 
  \renewcommand{\arraystretch}{1.2}
  \begin{tabular}{@{}lcccc@{}}
    \toprule
    Metric & ViTacTip & MagicTac & GelSight & GelSightWM \\
    \midrule
    Resolution [MP]   & 0.9  & 0.3  & 8.0  & 8.0  \\
    Thickness [mm]    & 50.0 & 15.0 & 4.25 & 4.25 \\
    FOV [mm$^{2}$]    & 1257 & 304  & 266  & 266  \\
    FPS [Hz]          & 30   & 60   & 25   & 25   \\
    \bottomrule
  \end{tabular}
    \vspace{-0.7cm}
\end{table}


\subsection{Force and Localization Calibration Error} 

\begin{table}[ht]
    \centering
    \captionsetup{font=small}
    \caption{Performance metrics comparison between sensors.}
    \begin{tabular}{lcccc}
        \toprule
        Sensor & Variable & MAE & $R^2$ & sMAPE \\
        \midrule
        ViTacTip   & $F_{xy}$ & \textbf{$\downarrow$0.0102} & 0.9617  & 0.6456 \\
                   & $F_z$    & \textbf{$\downarrow$0.0164} & 0.9775  & 0.1259 \\
        MagicTac   & $F_{xy}$ & 0.0498   & 0.9304  & 0.6081 \\
                   & $F_z$    & 0.0544   & 0.9675  & 0.1178 \\
        GelSight   & $F_{xy}$ & 0.0245   & \textbf{$\uparrow$0.985} & \textbf{$\downarrow$0.4815} \\
                   & $F_z$    & 0.0361   & 0.988  & 0.0809 \\
        GelSightWM & $F_{xy}$ & 0.0575   & 0.893  & 0.8525 \\
                   & $F_z$    & 0.0269   & \textbf{$\uparrow$0.991} & \textbf{$\downarrow$0.0651} \\
        \midrule
        ViTacTip   & $P_{xy}$ & 0.3514   & \textbf{$\uparrow$0.9948} & 0.1187 \\
                   & $P_z$    & 0.1781   & 0.9409  & 0.1668 \\
        MagicTac   & $P_{xy}$ & 0.2055   & 0.9925  & 0.1340 \\
                   & $P_z$    & 0.0516   & 0.9510  & 0.1065 \\
        GelSight   & $P_{xy}$ & 0.2479   & 0.983  & 0.1485 \\
                   & $P_z$    & 0.0325   & 0.9735 & 0.0821 \\
        GelSightWM & $P_{xy}$ & \textbf{$\downarrow$0.1445} & 0.9927 & \textbf{$\downarrow$0.0985} \\
                   & $P_z$    & \textbf{$\downarrow$0.0294} & \textbf{$\uparrow$0.9766} & \textbf{$\downarrow$0.0705} \\
        \bottomrule
    \end{tabular}
    \label{tab:metrics_comparison}
     \vspace{-0.2cm}
\end{table}

Table~\ref{tab:metrics_comparison} shows force and localization regression across three important metrics: MAE reflects absolute error in physical units, $R^2$ quantifies trend fidelity, and sMAPE emphasizes proportional error relative to signal magnitude. 

\textbf{Force regression.} ViTacTip attains the lowest MAE in both $F_{xy}$ and $F_z$, indicating the best absolute force accuracy. 
Its slightly higher sMAPE arises because its force range is smaller (up to $0.7$\,N), therefore identical absolute deviations constitute a larger fraction of signal magnitude. 
Design-wise, ViTacTip’s transparent elastomer with surface markers yields strong, trackable lateral displacements and high mechanical sensitivity per unit load, which reduces absolute error in both $F_{xy}$ and $F_z$. 
MagicTac’s internal grid supplies partial shear cues - improving $F_{xy}$ over a marker-free design - but depth-dependent refraction and multi-path effects increase variance, which is consistent with its higher $F_z$ MAE. 
GelSight remains balanced with consistently high $R^2$ ($>0.98$), indicating it closely tracks force trends even when absolute errors are not minimal.
GelSightWM performs well in $F_z$ (low MAE and sMAPE) but less well in $F_{xy}$: removing markers produces smoother, more monotonic photometric changes that support normal-force mapping while offering weaker tangential cues.

\textbf{Localization regression.} 
ViTacTip shows high $R^2$ in lateral positioning alongside larger MAEs in $P_{xy}$ and $P_z$, because the spherical optical path and dome compliance introduce non-linear warping and Rim-shadowing, i.e., optical artefacts at the sensor boundary where the dome blocks illumination and reduces edge accuracy, arising from the spherical optical path and dome compliance, causing self-occlusion and uneven lighting \cite{ward2018tactip}. This yields near-correct trends but an offset in absolute coordinates. 
MagicTac attains a competitive $P_{xy}$ but a weaker $P_z$, consistent with grid-induced refractive distortions that vary with indentation and confound depth estimation. 
GelSightWM yields the lowest MAE in both $P_{xy}$ and $P_z$ with high $R^2$ ($>0.97$). 
A likely cause is that its marker-less, reflective surface produces stable photometric gradients, allowing reliable localization without marker texture competing for features. 
GelSight behaves similarly, but its surface markers can slightly perturb the contact patch appearance, explaining modestly higher MAEs. 

The differences mainly come from the marker design. ViTacTip combines a soft elastomer with surface markers: the high compliance produces large, easily visualizable deformations, which explain its low MAE in force regression. However, its limited force range means that proportional accuracy (sMAPE) does not improve to the same extent, so its advantage is confined to absolute error metrics rather than overall scaling. GelSightWM uses the same elastomer as GelSight but without markers, generating a cleaner deformation image; this reduces shear sensitivity in $F_{xy}$ but benefits $F_z$ estimation and yields more reliable localization. MagicTac relies on an internal grid that bends and refracts under shear, giving it  better $F_{xy}$ performance than a marker-free design, but the grid itself bends inconsistently under load, introducing optical distortions and reducing stability. These design choices explain the trade-offs we observe in force and position regression.

\subsection{Spatial Resolution (\texorpdfstring{$SR(\varepsilon)$}{SR(epsilon)})}

By gradually tightening the spatial tolerance threshold, from broader error allowances to finer ones, we effectively characterize their spatial resolution as shown in Fig. \ref{fig:spatial-resolution} (c). When the spatial tolerance threshold \(\varepsilon\) is above 5 mm, all sensors exhibit near-perfect accuracy, with negligible differences. However, when the tolerance is tightened to 0.05 mm, performance differences become apparent. At this stricter level, ViTacTip’s accuracy drops sharply to 80.6\%, while the GelSight series still remains at 99.0\% and MagicTac at 98.3\%. These results demonstrate that the Gelsight series and Magictac sensor both offer exceptionally high spatial resolution, while ViTacTip’s performance deteriorates significantly under finer resolution demands.

The differences can be attributed to the mechanical and morphological properties of the elastomer layer. ViTacTip employs a soft, dome-shaped, thick gel (50 mm). This geometry, combined with its high compliance, allows it to conform well to object surfaces, but equally distributes strain and smooths out fine local features, leading to rapid accuracy degradation under stricter tolerance thresholds. By contrast, the GelSight series and MagicTac use stiffer, flatter, and thinner elastomer layers that keep strain confined more locally to the contact region, and maintain sharper indentation traces and thus support sharper spatial resolution.

In addition, ViTacTip employs a relatively low-resolution camera combined with a comparatively large sensing field-of-view, which reduces its effective pixel density on the contact surface. When the indenter feature size approaches this imaging limit, the model accuracy deteriorates more rapidly. In contrast, the GelSight series and MagicTac achieve a higher effective sampling density (pixels per mm$^2$) because their camera resolutions are higher relative to their smaller sensing FOV.

In conclusion, both material properties and morphology have a significant impact on the spatial resolution. Soft gels (e.g., PDMS, hydrogel) conform well to the contact surface, but excessive compliance - especially when combined with thick-layered, dome-shaped geometry - induces global strain that blurs local detail by its internal adhesion and shear. 
Stiffer, thin-layer designs preserve well-defined indentation patterns, though they require higher loads to conform fully. 
Therefore, intermediate-modulus transparent elastomers in flat geometries are often considered optimal, as they balance compliance and fidelity in feature capture. Finally, spatial resolution is jointly constrained by the imaging quality: a high effective pixel density (camera resolution relative to sensing field-of-view) is essential to sample these features once they are formed on the elastomer surface.





\subsection{Mechanical Sensitivity (\texorpdfstring{$S$}{S})}

\begin{figure}
    \centering
    \includegraphics[width=1\linewidth]{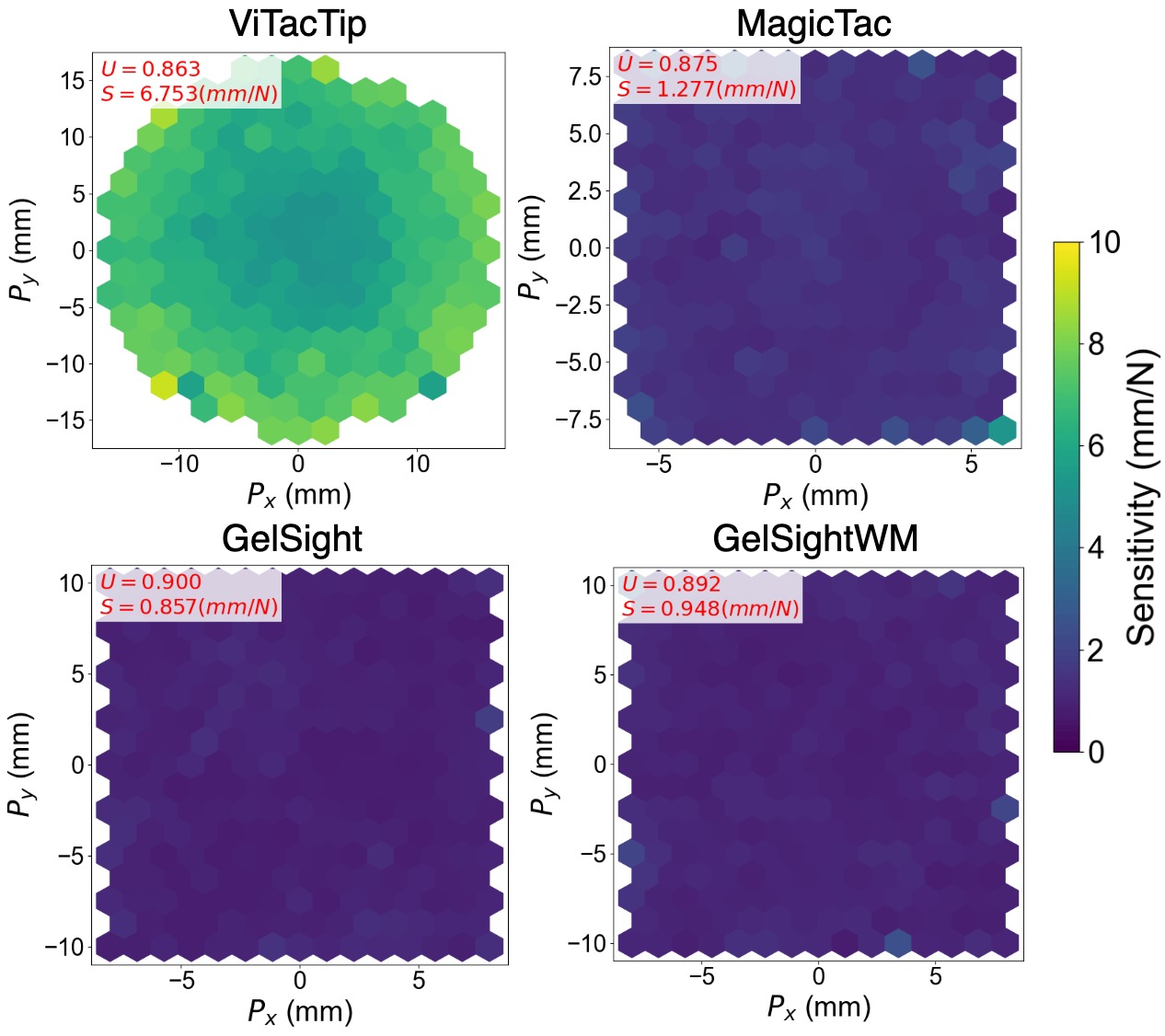}
     \captionsetup{font=footnotesize}
    \caption{Sensitivity heat map for MagicTac, GelSightWM, GelSight and ViTacTip, alongside their respective uniformity and average sensitivity values.}

                \captionsetup{font=small}
    \label{fig:Sensitivity}
\end{figure}

Fig. \ref{fig:Sensitivity} illustrates the mechanical sensitivity distribution on a logarithmic scale across the four VBTSs. MagicTac, GelSightWM and GelSight exhibit relatively uniform and low sensitivity levels, reflected by the predominantly blue color scale across their contact regions. In contrast, ViTacTip not only shows higher overall sensitivity values (shifted toward the green–yellow range) but also a pronounced spatial variation: the central region maintains a lower sensitivity of (\(S\!\approx\!5\)), whereas the edges approach nearly twice this value.
This indicates that, for the same applied force, \textbf{ViTacTip} deforms more than the other sensors, especially near the periphery, consequently exhibiting higher effective compliance and stronger boundary effects. For any given \(F_z\), it therefore exhibits a larger displacement \(P_z\), but also a reduced usable force range. However, this higher compliance comes at the cost of reduced 
uniformity (\(U\)), meaning that predictions of force or displacement 
may vary considerably depending on contact location. In contrast, the 
stiffer gels of GelSight and MagicTac maintain higher \(U\) values, despite 
lower \(S\) values, suggesting they are better suited for tasks requiring consistent calibration across the surface. In summary, ViTacTip favors sensitivity at the expense of spatial consistency, whereas GelSight-type sensors emphasize stability over the range of deformation.

\subsection{Spatial Robustness (\texorpdfstring{$R_{\text{spatial}}$}{Rspatial})}

\begin{table}
  \centering
  \caption{Overall spatial robustness variability (mean STD of distance and depth conditions).}
  \label{tab:spatial_std_overall}
  \small
  \setlength{\tabcolsep}{6pt}
  \renewcommand{\arraystretch}{1.1}
  \begin{tabular}{lcccc}
    \toprule
    Metric & GelSight & GelSightWM & ViTacTip & MagicTac \\
    \midrule
    $P_{xy}$ & 0.484 & 0.300 & \textbf{$\downarrow$0.258} & 0.266 \\
    $P_{z}$  & \textbf{$\downarrow$0.028} & \textbf{$\downarrow$0.027} & 0.144 & 0.043 \\
    $F_{xy}$ & 0.032 & 0.061 & \textbf{$\downarrow$0.009} & 0.069 \\
    $F_{z}$  & 0.109 & 0.147 & \textbf{$\downarrow$0.025} & 0.125 \\
    \bottomrule
  \end{tabular}
  \vspace{2pt}

  \footnotesize\textit{Note: values are mean STDs for distance and depth conditions. Position metrics ($P_{xy}$, $P_{z}$) in mm; force metrics ($F_{xy}$, $F_{z}$) in N. The lower the value, the better the performance.}
    \vspace{-0.5cm}
\end{table}

Fig.~\ref{fig:Spatial robustness} shows that spatial robustness is clearly both channel and design-dependent, rather than being a uniform edge effect. Edge degradation is strong in the GelSight-series for $P_{xy}$,$F_{xy}$, and in all sensors for $F_{z}$, while ViTacTip largely resists such trends in the force domain. Depth profiles also diverge: $P_{xy}$ errors drop sharply after shallow contacts, $P_{z}$ curves are generally flat with occasional U-shapes, and both force channels grow with indentation depth, though much less so for ViTacTip.

A summary of the experimental results $P_{xy}$, $P_z$, $F_{xy}$, $F_z$ for the four representative sensors is outlined below:

\noindent\textbf{Lateral Position ($P_{xy}$):} Across all sensors, localization error increases towards the periphery (Fig.~\ref{fig:Spatial robustness}(a)). ViTacTip attains the lowest overall MAE ($\sim$0.5\,mm) and variability (STD $\sim$0.25\,mm), reflecting stable localization afforded by its spherical cavity and uniform imaging geometry. In contrast, GelSight and GelSightWM exhibit clear error inflation beyond radius $>0.8$ (up to 1.5 mm), consistent with peripheral optical distortion from planar imaging. Normalized-depth analysis (Fig.~\ref{fig:Spatial robustness}(b)) indicates that the decline in errors arises because shallow contacts create incomplete contact areas that give ambiguous lateral cues, while deeper indentations enlarge and stabilize the contact patch, improving localization.

\noindent\textbf{Normal Position ($P_{z}$)}:
Depth-estimation errors remain low ($<0.25\,mm$), increasing only modestly towards the periphery. GelSight-based sensors provide the most stable performance (STD $\sim 0.027\,mm$), because of a thinner, stiffer gel that provides clearer depth cues, while MagicTac shows comparable performance, likewise, due to its thin gel.
In contrast, ViTacTip shows much higher peripheral variance ($\sim 0.13\,mm$), likely due to rim shadowing. Both ViTacTip and MagicTac exhibit a mild U-shaped trend with depth. Errors are greater when contact is shallow  (insufficient compression for reliable photometric cues) and, equally, when indentations are very deep (optical distortion near the gel boundary).

\noindent\textbf{Lateral Shear Force ($F_{xy}$)}:
This performance metric depends on marker-displacement visibility. ViTacTip outperforms the other sensors, maintaining low error and variance across the surface. In contrast, GelSightWM ($0.061$ N) suffers from the absence of markers, while MagicTac’s internal grid introduces depth-dependent refraction that destabilizes feature tracking. Comparison analysis against normalized depth further reveals that MAE increases monotonically with indentation depth for all GelSight-based sensors, indicating that shear-force estimation becomes progressively more difficult as marker deformations intensify.

\noindent\textbf{Normal Force ($F_{z}$)}:
ViTacTip again achieves the lowest normal-force MAE ($\sim$0.02\,N), maintaining stable performance across both spatial dimensions. The other three sensors show substantial error growth toward the edges as well as with increasing depth, most likely due to boundary stiffness gradients in planar elastomers and reduced uniformity of vertical marker motion near the edges. This highlights a fundamental limitation of flat-gel architectures for normal force estimation.

In summary, position-localization errors ($P_{xy}$, $P_z$) are driven primarily by optical and geometric distortions rather than marker density. In contrast, the accuracy of the force estimation ($F_{xy}$, $F_z$) depends critically on the visibility of the marker and the uniformity of the displacement of the marker, with the geometry of the sensor and the stiffness of the edges also playing a substantial role. ViTacTip’s spherical geometry and dense, uniform marker pattern deliver high accuracy and stable performance in force estimation, whereas planar GelSight-based sensors exhibit spatial degradation, particularly toward the periphery and under deeper indentations.


\begin{figure*}[t]
    \centering
    \includegraphics[width=1\linewidth]{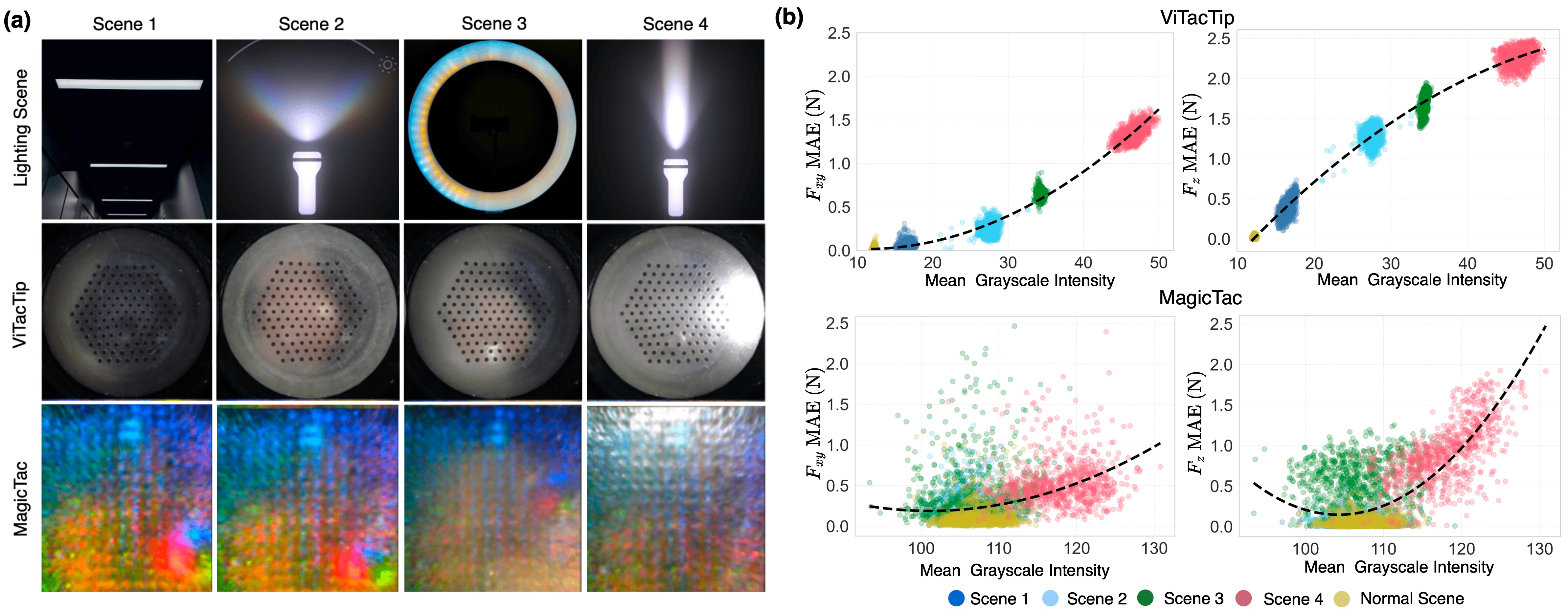}
    \captionsetup{font=footnotesize}
    \caption{\textbf{(a) Lighting scenes vs visual output of ViTacTip and MagicTac.} Four different lighting scenes, notably dim room lighting, diffuse flash lighting, streamer ring lighting, and streamed flash lighting, were tested to evaluate their effect on the sensor's visual output. Sample frames are shown from ViTacTip (top) and MagicTac (bottom) under four illumination conditions.
\textbf{(b) Force-regression MAE plotted against mean grayscale intensity:} ViTacTip shows increasing error with brightness, while MagicTac stays comparatively stable.}
    \label{fig:light robustness}
    \vspace{-0.4cm}
\end{figure*}


\subsection{Lighting Robustness (\texorpdfstring{$R_{\text{light}}$}{Rlight})}

For this metric, we restrict evaluation to MFM-based sensors, which provide multi-modal sensing but are susceptible to external light sources. We created new test data by seting up four lighting scenes (in Fig. \ref{fig:light robustness}(a)) that are different from the illumination condition of the dataset used in calibration. As shown in the raw visual output under the different lighting scene, the ViTacTip sensor is highly susceptible to changes in lighting intensity due to its material properties and geometry. 

In Fig.\ref{fig:light robustness}(b), the maximum light intensity in the experiments was up to approximately 7 times greater than that experienced during training conditions. In contrast, MagicTac exhibits greater resistance to external light noise due to the refraction effects caused by its internal grid structure. As a result, diffuse light sources (scene 1~3) had minimal impact on its average light intensity, with only point light sources causing noticeable fluctuations. This suggests that MagicTac effectively mitigates the influence of occlusions on illumination. Although MagicTac’s physical design offers a degree of light shielding (resulting in small intensity deviations), the prediction error variance increases markedly under the same lighting condition, resulting in poorer model robustness than ViTacTip despite stable raw intensities. Under point light conditions (scene 4), ViTacTip’s MAE increases by a factor of about ten relative to its performance under normal lighting scene, even though the light intensity is only about seven times higher. In contrast, with MagicTac, the intensity-induced error is much lower, but MAE values have high variance, which likely arises from its lattice-embedded markers. Variations in lighting make it harder to align the captured grid with the training reference, reducing marker distinctiveness and impairing tracking. Furthermore, MagicTac leverages photometric stereo principles by utilizing red, green, and blue light to enhance z-axis depth and force regression. When external illumination interferes with the internally generated light, errors inherent to the photometric stereo method may also arise. 

\begin{table}[t]
\centering
\captionsetup{font=small}
\caption{Light robustness comparison between ViTacTip (left) and MagicTac (right)}
\resizebox{\linewidth}{!}{%
\begin{tabular}{lcccc}
\toprule
Scene & $F_{xy}$ & $F_{z}$ & $P_{xy}$  & $P_{z}$ \\
\midrule
S1 & 11.5\% / 1.8\% & 1.2\% / 35.1\% & 2.0\% / 2.7\% & 2.0\% / 4.7\% \\
S2 & 17.6\% / 1.2\% & 1.4\% / 2.8\%  & 3.9\% / 1.1\% & 3.3\% / 4.2\% \\
S3 & 4.4\% / 0.1\%  & 1.5\% / 0.0\%  & 6.5\% / 0.0\% & 3.2\% / 0.0\% \\
S4 & 3.4\% / 2.3\%  & 1.7\% / 0.6\%  & 7.3\% / 1.0\% & 3.6\% / 1.4\% \\
\midrule
Mean & \textbf{$\uparrow$9.3\%} / 1.4\% & 1.5\% / \textbf{$\uparrow$9.6\%} & \textbf{$\uparrow$5.0\%} / 1.2\% & \textbf{$\uparrow$3.0\%} / 2.6\% \\
\bottomrule
\end{tabular}%
}
\vspace{-0.4cm}
\label{tab:lighting}
\end{table}

In summary, ViTacTip demonstrates higher robustness for tangential force and position estimation, while MagicTac shows superior robustness in normal force estimation. For normal positions, the two sensors perform comparably, with ViTacTip showing a slight overall advantage.



\begin{figure*}[t]
    \centering
    \includegraphics[width=0.9\linewidth]{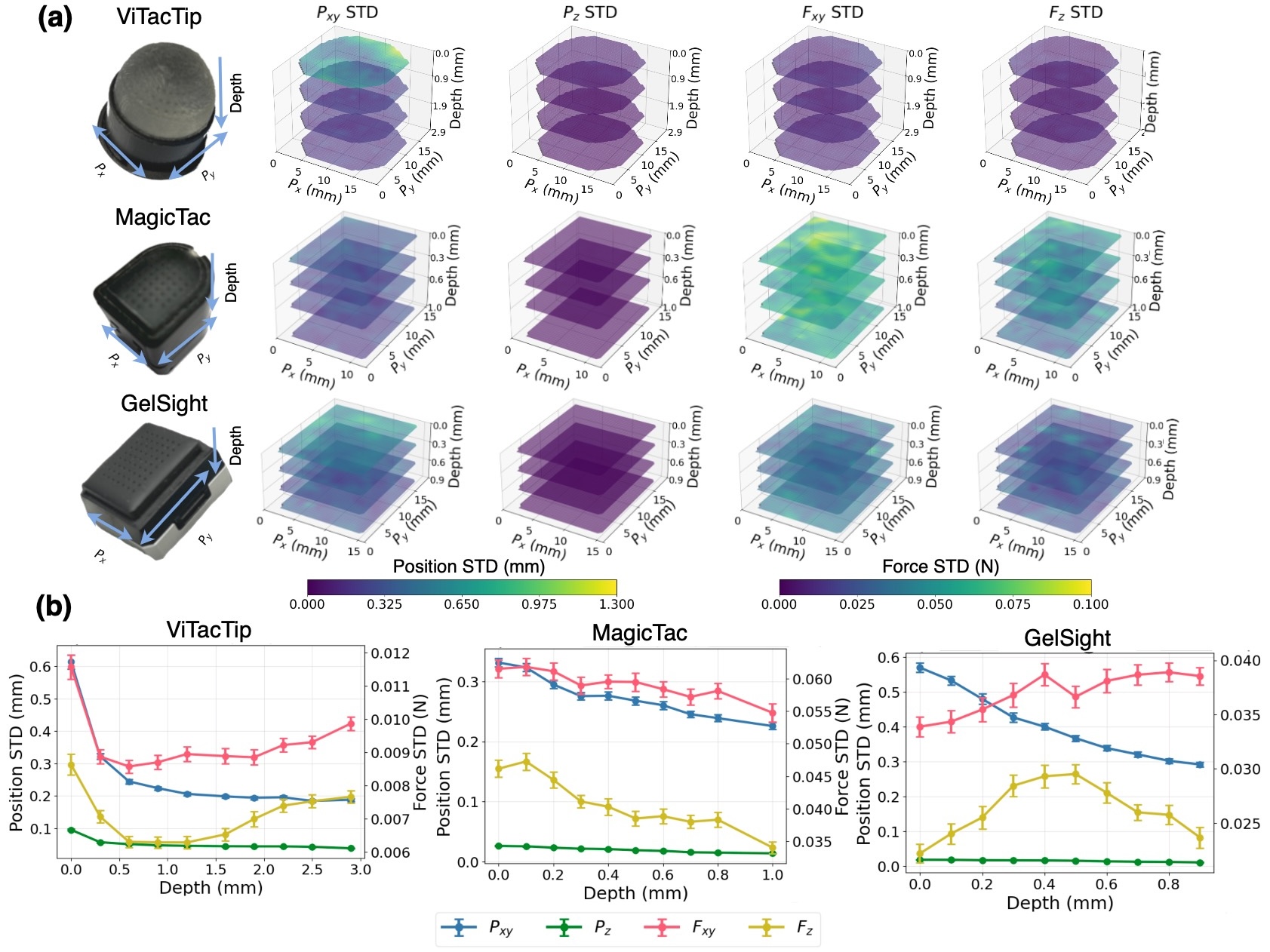}
    \captionsetup{font=footnotesize}
        \caption{Repeatability analysis for ViTacTip, MagicTac, and Gelsight. (a) Standard deviation (STD) of predicted force components ($F_{xy}$, $F_z$) and positional components ($P_{xy}$, $P_z$) across multiple trials at various depths of VBTS.   (b) Line plots of repeatability (STD) of the predicted outputs as a function of indentation depth. Positional STDs ($P_{xy}$,$P_z$, left y‑axis, in mm) and force STDs ($F_{xy}$, $F_z$, right y‑axis, in N) are shown for each sensor.}
\vspace{-0.5cm}
        \label{fig:Repeatability}
\end{figure*}


\subsection{Repeatability (\texorpdfstring{${Rep}$}{Rep})}

We used the ResNet-based model calibrated for force and contact-localization regression as predictors.  Fig.~\ref{fig:Repeatability}(a) shows predictions from repeated trials across different indentation planes, and Fig.~\ref{fig:Repeatability}(b) illustrates the mean STD curves as a function of indentation depth. These depth-dependent plots illustrate how repeatability varies, while the aggregated $\mathrm{Rep}_c$ values in Table~\ref{tab:repeatability} provide a global comparison. 

For \textbf{GelSightWM}, repeatability under shear-dominant stimuli is not reported because its marker-less, intensity-mapping design provides weak observability of tangential components; accordingly, we focus on sensors with explicit displacement cues for shear-dominant stimuli. 


For the predicted force variables ($F_{xy}$, $F_z$), ViTacTip consistently demonstrates the most stable force predictions across various depths (lowest STDs; $F_{xy}$: 0.006, $F_z$: 0.007). The next best is GelSight ($F_{xy}$: 0.025, $F_z$: 0.026) followed by MagicTac ($F_{xy}$: 0.041, $F_z$: 0.040). Primarily attributable to  ViTacTip’s soft dome shape, it produces large, smooth deformations under load, which are easier for the model to regress consistently. MagicTac varies more because even small differences in repetitive presses can easily change the internal grid's structure, which the model interprets
as force changes.

In terms of predicted positional variables ($P_{xy}$, $P_z$),  ViTacTip again demonstrates the best repeatability ($P_{xy}$: 0.166), though it shows higher variability at very shallow contacts due to unstable skin deformation under minimal loading.  GelSight achieves the most stable depth localization ($P_z$: 0.015), but performs poorly in lateral localization ($P_{xy}$: 0.278). This suggests that GelSight’s opaque elastomer geometry, rather than its surface markers, governs its positional repeatability - indeed, the dense marker layer may add noise rather than provide useful cues for localization. MagicTac shows intermediate repeatability ($P_{xy}$: 0.188, $P_z$: 0.020), performing better than GelSight laterally, but worse axially.

Overall, ViTacTip performs best on repeatability, benefiting from its compliant dome and optical transparency. Variability is only notable during the earliest shallow contacts, when the visible cues are too subtle. GelSight is particularly stable in depth localization, but its lateral repeatability is limited. The opaque elastomer and dense marker distribution appear to constrain reliable $P_{xy}$ tracking. While MagicTac provides intermediate positional repeatability, its force estimates remain the most variable, as the internal grid marker design is prone to instability under repeated loading. Notably, ViTacTip was operated over a wider indentation span (3.5 mm) but a narrower force range (0.7 N), whereas the others covered shallower depths (1 mm) with higher forces (beyond 2 N). This indicates that its soft and compliant dome converts small forces into large, smooth deformations, producing highly stable cues under low loading. By contrast, stiffer designs deform less but tolerate larger forces, so their repeatability is reduced even though they can cope with stronger contacts. This suggests that repeatability is inherently linked to material compliance, and therefore, that sensors designed for stable low-force interaction should favor softer, thicker elastomers, while stiffer gels would be better suited to tasks requiring force endurance.

\begin{table}[t]
\centering
\captionsetup{font=small}
\caption{Repeatability results for position and force variables}
\label{tab:repeatability}
\small
\begin{tabular}{lcccc}
\toprule
Sensor & $P_{xy}$ & $P_{z}$ & $F_{xy}$ & $F_{z}$ \\
\midrule
ViTacTip  & \textbf{$\downarrow$0.166} & 0.049 & \textbf{$\downarrow$0.006} & \textbf{$\downarrow$0.007} \\
GelSight  & 0.278 & \textbf{$\downarrow$0.015} & 0.025 & 0.026 \\
MagicTac  & 0.188 & 0.020 & 0.041 & 0.040 \\
\bottomrule
\end{tabular}
\vspace{-0.4cm}
\end{table}

\subsection{Sensor Evaluation and Decision Guide}

\begin{figure*}[t]
    \centering
    \includegraphics[width=0.95\linewidth]{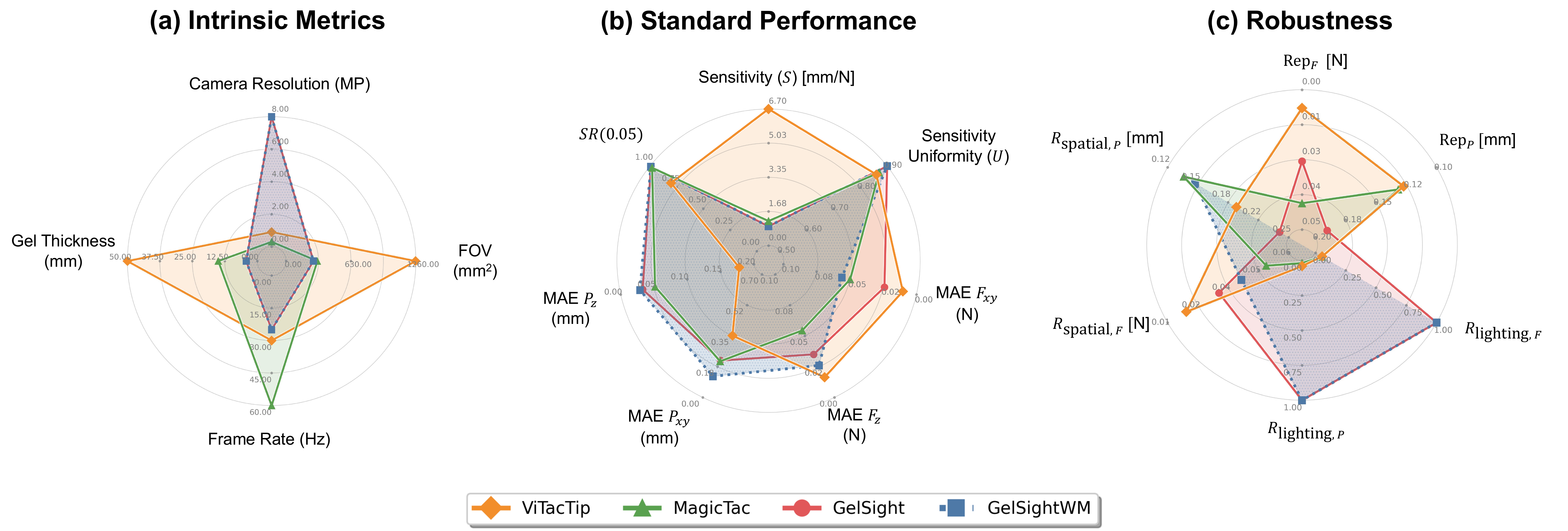}
 \captionsetup{font=footnotesize}
    \caption{Radar plots comparing the four VBTSs on three evaluation themes: (a) Intrinsic properties (camera resolution, field of view, frame rate, gel thickness and sensitivity); (b) Standard performance (spatial resolution and mean absolute errors in force and position, and sensitivity); (c) Robustness (including repeatability, spatial robustness, and light robustness in both force and position metrics). Larger areas correspond to stronger performance across the evaluated metrics.}
        \label{fig:radar_summary}
    \vspace{-0.7cm}
\end{figure*}

Fig.~\ref{fig:radar_summary} compares the four VBTSs on three categories: intrinsic properties, standard performance, and robustness. The radar plots' numerical values are taken from the corresponding quantitative tables and figures. Specifically, for radar plot (a), values are extracted from Table~\ref{tab:vbts_specs}. For radar plot (b), the MAE and related performance metrics are taken from Table~\ref{tab:metrics_comparison}, in which the spatial resolution corresponds to Fig.~\ref{fig:spatial-resolution}(d) with tolerance set to $0.05$\,mm, and sensitivity values are given in Fig.~\ref{fig:Sensitivity}. For radar plot (c), lighting robustness values are taken from Table~\ref{tab:lighting}, while spatial robustness and repeatability are taken from Tables~\ref{tab:spatial_std_overall} and Table~\ref{tab:repeatability}, respectively. As the original tables report $F_{xy}$ and $F_{z}$ separately, their mean is computed to yield the ``Force'' axis in the radar chart; similarly, $P_{xy}$ and $P_{z}$ are averaged out to obtain the ``Position'' axis.  

Intrinsic properties are related to the hardware specifications of the sensors. These metrics reflect the fundamental design trade-offs imposed by the sensor’s hardware and optical pathways. Standard performance characterizes the sensor’s accuracy and resolution under standard conditions. This includes regression performance (e.g., force/pose estimation error), compliance of the elastomer, and spatial resolution for detecting fine contact patterns. These indicators capture the sensor’s “nominal” capability when operating without external perturbations. Robustness is a measure of how consistently the sensors maintain their performance under varying conditions. The table includes repeatability and spatial robustness, quantified using the standard deviation of model predictions to reflect how stable the output remains across trials and positions. Lighting robustness is also reported as a dimensionless parameter, capturing the sensitivity of each sensor to illumination changes. These results reveal distinct trade-offs across categories, indicating different application niches for each sensor type.

\textbf{ViTacTip} achieves relatively strong force accuracy and the best repeatability, reflecting highly stable behavior across different stages. However, its optical path makes it sensitive to external illumination, and its depth estimation (\(P_z\)) and spatial resolution are weaker. Materially, it has the softest gel, the greatest thickness, and the largest field of view. These features make it well-suited to tasks emphasizing force estimation on soft or deep contacts, where fine texture discrimination and sub-millimeter localization are less critical.

\textbf{MagicTac} excels in contact localization accuracy and uniformity, with spatial resolution only slightly behind the GelSight series, while also offering the highest frame rate. In contrast, its force estimation is generally weaker, and its illumination robustness is poor - likely a consequence of the internal grid producing refractive artifacts under varying light conditions. This profile favors fast, planar localization in scenes where force accuracy demands are modest and illumination can be controlled.

\textbf{GelSight} offers the highest camera resolution and the most uniform gel sensitivity, which is beneficial to fine-grained contact localization. These advantages come with trade-offs: the gel is comparatively stiff (lower sensitivity), and the camera’s frame rate is modest. Edge-area distortion is also evident in the repeatability panel. Nevertheless, its force response is relatively uniform across the field, yielding a well-rounded overall profile.

\textbf{GelSightWM} delivers stable performance in normal force and depth estimation (\(F_z/P_z\)) but remains weaker in shear force (\(F_{xy}\)). Its planar contact localization (\(P_{xy}\)) is competitive among the four; the absence of surface markers likely encourages the model to rely more directly on contact-induced cues rather than visual texture, which benefits contact detection in the plane.

For deep or soft contacts, where normal-force sensitivity is central, \textbf{ViTacTip} is the obvious choice despite its light sensitivity. Shallow or firm contacts benefit from \textbf{GelSight} or \textbf{MagicTac}, with GelSight preferred for depth (\(P_z\)) and MagicTac for planar localization (\(P_{xy}\)) when speed matters. Applications that stress accurate shear force should still prioritize \textbf{ViTacTip}, but when robustness to illumination is paramount, \textbf{GelSight}/\textbf{GelSightWM} are safer defaults. In scenarios where shear is not required, \textbf{GelSightWM} provides a practical and sufficiently stable solution.


\section{Conclusions and Future Work}

In this work, we introduced TacEva, a systematic evaluation framework for VBTSs. TacEva defines a comprehensive set of standardized performance metrics, provides structured and reproducible evaluation pipelines, and demonstrates their application across multiple sensors with distinct sensing mechanisms. TacEva delivers the first unified framework for consistent and quantitative assessment of VBTS designs, enabling direct comparison and performance-guided insights into sensor optimization.

Looking ahead, the next step is to extend TacEva into a dedicated, community-maintained framework that catalogs a broad range of VBTSs with their core attributes, while also hosting benchmark datasets and standardized protocols for representative tasks. Furthermore, the scope of evaluation should be expanded to encompass task-level and system-level performance metrics. Collectively, these developments would establish TacEva as a comprehensive platform that integrates shared resources with application-oriented evaluation protocols, thereby enabling objective comparison, informed sensor selection, and performance-driven innovation for the next generation of VBTSs.

\section*{Acknowledgement}

The authors would like to thank Mish Toszeghi (Queen Mary University of London) for helpful comments and proofreading assistance.
\section*{Conflict of Interest}
The authors declare no conflict of interest.

\bibliographystyle{IEEEtran}

\bibliography{IEEEabrv,ref}

\end{document}